\documentclass[sigconf]{acmart}
\usepackage{ulem}

\AtBeginDocument{%
  \providecommand\BibTeX{{%
    \normalfont B\kern-0.5em{\scshape i\kern-0.25em b}\kern-0.8em\TeX}}}

\acmSubmissionID{186}
\usepackage{multirow}
\usepackage{textcomp} 
\usepackage{color}
\definecolor{DarkGreen}{rgb}{0.0, 0.5, 0.0}
\definecolor{LimeGreen}{rgb}{0.3, 0.5, 0.2}

\begin{document}

\title{AnimateLCM: Computation-Efficient Personalized Style Video Generation without Personalized Video Data}

\author{Fu-Yun Wang}
\email{fywang@link.cuhk.edu.hk}
\affiliation{%
  \institution{MMLab, CUHK}
  \country{Hong Kong SAR}
}
\author{Zhaoyang Huang}
\email{zhaoyanghuang@avolutionai.com}
\affiliation{%
  \institution{Avolution AI}
  \country{China}
}
\author{WeiKang Bian}
\email{wkbian@outlook.com}
\affiliation{%
  \institution{MMLab, CUHK}
  \country{Hong Kong SAR}
}
\author{Xiaoyu Shi}
\email{xiaoyushi@link.cuhk.edu.hk}
\affiliation{%
  \institution{MMLab, CUHK}
  \country{Hong Kong SAR}
}
\author{Keqiang Sun}
\email{kqsun@link.cuhk.edu.hk}
\affiliation{%
  \institution{MMLab, CUHK}
  \country{Hong Kong SAR}
}

\author{Guanglu Song}
\email{guanglusong@foxmail.com}
\affiliation{%
  \institution{SenseTime Research}
  \country{China}
}
\author{Yu Liu}
\email{liuyuisanai@gmail.com}
\affiliation{%
  \institution{SenseTime Research}
  \country{China}
}
\author{Hongsheng Li}
\email{hsli@ee.cuhk.edu.hk}
\affiliation{%
  \institution{MMLab, CUHK}
  \institution{Centre for Perceptual and Interactive
Intelligence (CPII)}
  \country{Hong Kong SAR}
}

\renewcommand{\shortauthors}{Fu-Yun Wang, et al.}

\begin{abstract}
This paper introduces an effective method for computation-efficient personalized style video generation without requiring access to any personalized video data. \textit{It reduces the necessary generation time of similarly sized video diffusion models from 25 seconds to around 1 second while maintaining the same level of performance.} The method's effectiveness lies in its dual-level decoupling learning approach: 1) separating the learning of video style from video generation acceleration, which allows for personalized style video generation without any personalized style video data, and 2) separating the acceleration of image generation from the acceleration of video motion generation, enhancing training efficiency and mitigating the negative effects of low-quality video data.
\end{abstract}



\maketitle

\section{Introduction}
\label{sec:intro}
\begin{figure}[!h]
    \centering
    \includegraphics[width=0.85\linewidth]{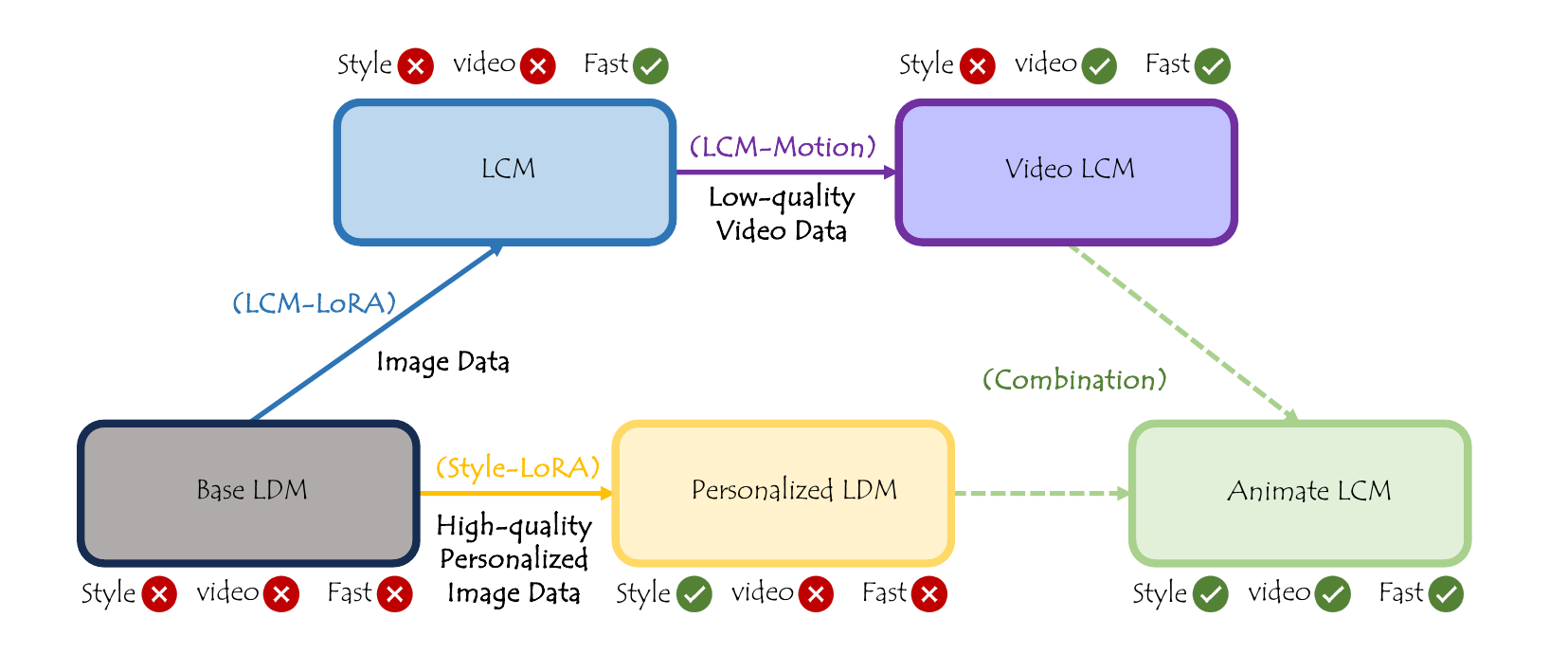}
\vspace{-1em}
    \caption{High-level overview of the pipeline of AnimateLCM.  {\textcolor{orange}{1) Fine-tune the base LDM on the high-quality personalized style image data for stylized image generation.} \textcolor{blue}{2) Accelerate the base LDM into LCM for fast image generation.} \textcolor{violet}{3) Accelerate and extend the LCM into video LCM for fast video generation.}  \textcolor{LimeGreen}{4) Combine the weights of personalized LDM and video LCM into AnimateLCM for computation-efficient personalized style video generation without any personalized video data.}} }
    \label{fig:overview}
\end{figure}

Over the past few years, the field of video generation has made significant strides, thanks to the utilization of video diffusion models~\cite{ho2022video,make-a-video,shi2024motion}. Currently, commonly applied video diffusion models can generate short video clips of about 2 seconds with relatively high-quality and reasonable motions. Nevertheless, those video generation models still have two significant shortcomings: 

\begin{enumerate}
    \item \textbf{Slow generation speed.} The high-quality generation achieved by the diffusion model relies on the iterative denoising process that gradually transforms high-dimensional noises into real data. However, the nature of iterative sampling leads to slow generation and high computational burdens of the diffusion model whose generation is much slower than other generative models~(e.g., GAN)~\cite{yu2023magvit, gan}.  For example, even testing on a high-performance GPU A100, it still takes 25 seconds to generate a 2-second short video clips in 512p$\times$512p.
    \item \textbf{Inflexibility of generation style}. In general, the quality of video data is inferior to that of image data, and accurately annotating video data with textual information is more challenging. Consequently, high-quality video data is difficult to obtain. Using low-quality video data typically results in suboptimal generation outcomes. Furthermore, users tend to prefer generating videos with higher quality and diverse styles, such as 2D animation, 3D animation, ink painting, etc. However, collecting high-quality videos in these styles is often very difficult.
\end{enumerate}
Our approach effectively addresses the aforementioned issues without requiring complex steps. The core of our method lies in independently solving the problems of style learning and video generation acceleration, and then integrating them through weight fusion. By doing so, we only need to collect high-quality image data of specific styles for content learning, while utilizing lower-quality video datasets to learn motion characteristics and accelerate video generation. Additionally, it is worth noting that a video can essentially be regarded as a series of images over time, connected through motion relationships. Therefore, we further decouple the acceleration of video generation into two parts: the generation acceleration of images and the generation acceleration of video motion. Our experimental results demonstrate that this decoupled acceleration method significantly enhances training efficiency. We illustrates the high-level idea of our methods in Fig.~\ref{fig:overview}.

\section{Related Works}
\label{sec:relatedwork}
Diffusion Models
have gradually dominated the filed of image and video generation, though suffering from low generation speed. LCM-LoRA~\cite{cm,lcm,wang2024phased,geng2024consistency}, working as a versatile acceleration module for image  diffusion models, attracted huge attention. This work explores an versatile module, enabling the off-the-shell image diffusion models for computation-efficient personalized style video generation. 

\begin{figure}[t]
    \centering
\includegraphics[width=0.9\linewidth, trim=0 50 0 0, clip]{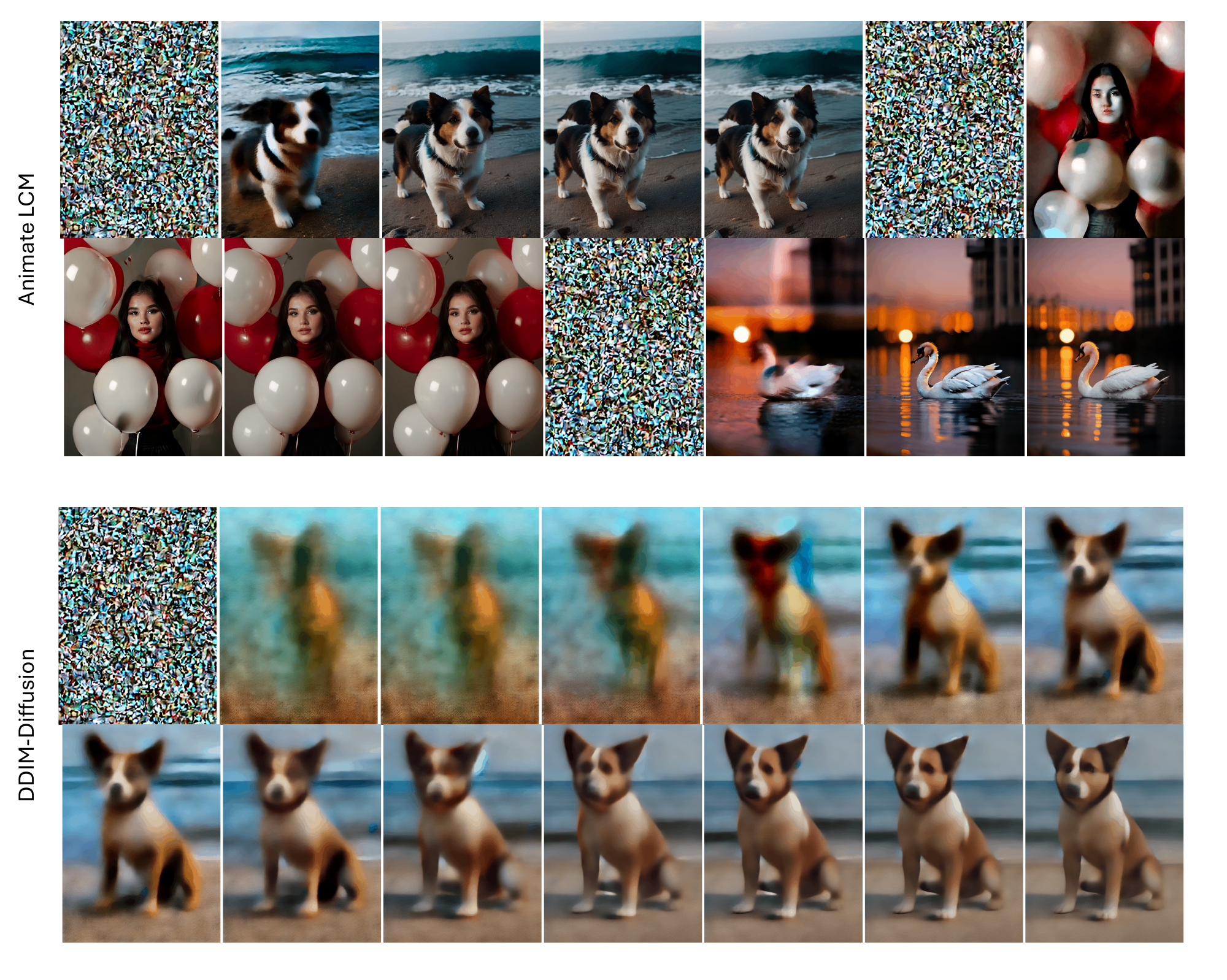}
\vspace{-1em}
    \caption{In the given denoising time budget, our model completes three high-quality generations, while video diffusion models are still in the process of denoising.}
    \label{fig:denoise}
\end{figure}

\section{Method}\label{sec:method}
Our model supports high-quality personalized style video generation without learning from any personalized video data. It also reduces the generation time by around 10--25 times compared to similarly sized diffusion models. Its effectiveness benefits from its dual-level decoupled learning strategy: 1) separating video style learning from generation acceleration, and 2) separating image generation acceleration and video motion generation acceleration. 
\subsection{Decoupling Style Learning and Acceleration}

\noindent \textbf{Fine-tuning base LDM on a personalized image dataset.} 
The base LDM is trained on a vast amount of text-image pairs that have not been thoroughly filtered. It can accept text inputs and generate corresponding images. Due to issues such as data quality and model capacity, this base model often struggles to accurately generate images that match the style described by the text. Fortunately, this pretrained base model has a good capability for fine-tuning. Typically, individuals can collect a few hundred or more private data samples to re-fine-tune the model, transforming the base LDM into a personalized LDM that can generate high-quality images in a specified style. Generally, since personal users have limited training resources, such as GPUs, they often adopt parameter-efficient fine-tuning methods, with LoRA~\cite{hu2021lora} being the most widely used. Specifically, the model's weight update can be expressed as \( w = w_0 + A B \), where \( w_0 \in \mathbb{R}^{d \times k} \) is the original weight of the model, \( A \in \mathbb{R}^{d \times r} \), and \( B \in \mathbb{R}^{r \times k} \), with \( r \ll \min(d, k) \). We can denote the $AB$ as $\tau_{personalize}$, functioning as a specific weight residual for stylized generation.

\noindent \textbf{AnimateLCM as a universal efficient video generation module.} 
Our motivation is that the process of accelerating the model through consistency distillation can still be seen as a fine-tuning process of the pretrained model. Therefore, the distillation acceleration process of the base LDM can still be viewed as learning a weight residual for the base LDM. Specifically, \( w_{\text{accelerated}} = w_{0} + \tau_{\text{accelerated}} \), where \( w_{\text{accelerated}}, w_{0}, \tau_{\text{accelerated}} \in \mathbb{R}^{d\times k} \). In this way, we obtain two weight residuals, \( \tau_{\text{accelerated}} \) and \( \tau_{\text{personalized}} \). We can linearly combine these residuals with the original weights for joint functionality. 
In practice, we use scaling factors \( \alpha \) and \( \beta \) to control the influence of different weight residuals, combining them as \( w_{\text{combined}} = w_0 + \alpha \tau_{\text{accelerated}} + \beta \tau_{\text{personalized}} \). It’s important to note that since these residuals are directly integrated with the original weights, they do not affect the actual computation speed.

\noindent \textbf{The decoupling learning approach eliminates the need for high-quality personalized style video data collection.} Overall, in the process described above, the stylized weight parameters are fine-tuned using a high-quality image dataset, while the weight residuals for acceleration are trained on general images and lower-quality video datasets, since high-quality stylized videos are hard to obtain. This approach allows us to combine the advantages of both methods, thereby eliminating the need for high-quality stylized video collections.

\subsection{Decoupling Image and Video Acceleration}
Videos can generally be viewed as sequences of images over time, with motion relationships between temporally adjacent frames. With this in mind, our motivation is that the acceleration weight residual mentioned earlier can be decomposed into two parts: one for learning the acceleration residuals in image generation, and the other for video motion generation. On one hand, learning from image data is typically easier and less costly than learning from video data. That is,
\begin{align}
    \tau_{accelerated} = \tau_{accelerated}^{image} + \tau_{accelerated}^{video} \, ,
\end{align}
where $\tau_{accelerated}^{image}$ aims to image generation acceleration while $\tau_{accelerated}^{video}$ aims to video motion generation acceleration.
On the other hand, the content of a video forms the basis for its motion; without clear spatial content, any temporal relationships become meaningless. Therefore, we propose first accelerating the base LDM for image generation to obtain the base LCM. From there, we extend the base LCM to accept video inputs and continue acceleration training on readily available low-quality video datasets. We found that this approach significantly speeds up the training process. In practice, we implement the $v_{accelerated}^{image}$ as the LoRA and implement the $v_{accelerated}^{video}$ as the motion module composed of temporal attention blocks. 

Thereby, the final weight is written as  
\begin{align}
    w' = w_0 + \alpha \tau_{personalized} + \beta \tau_{accelerated}^{image} + \gamma \tau_{accelerated}^{video}\, ,
\end{align}
where $\alpha, \beta, \gamma$ are all scaling factor. In practice, we find we generally have to set $\gamma = 1$ considering that the it is the only weight enables the video generation ability of base LDM. For $\alpha$ and $\beta$, users can scaling them to control the impact of different weight residuals. 

\section{Experiments}

\subsection{Benchmarks.} To evaluate our approach, we follow previous works, utilizing the widely used UCF-101~\cite{soomro2012ucf101} for validation. For each category, we generate $24$ videos with 16 frames in resolution $512 \times 512$ and thus generate $24 \times 101$ videos in total. We apply FVD~\cite{fvd} and CLIPSIM~\cite{hessel2021clipscore} as the validation metric.  For CLIPSIM, we rely on the CLIP ViT-H/14 LAION-2B~\cite{clip} to compute the mean value of the similarities of the brief caption and all the frames in the video. Following the validation choice in LCM~\cite{lcm}, we compare AnimateLCM with the teacher model using the DDIM~\cite{ddim} and DPM-Solver++~\cite{dpmsolver}. 

\subsection{Experimental Results}

\begin{figure}[t]
    \centering
\includegraphics[width=0.8\linewidth,trim=0 750 0 0, clip]{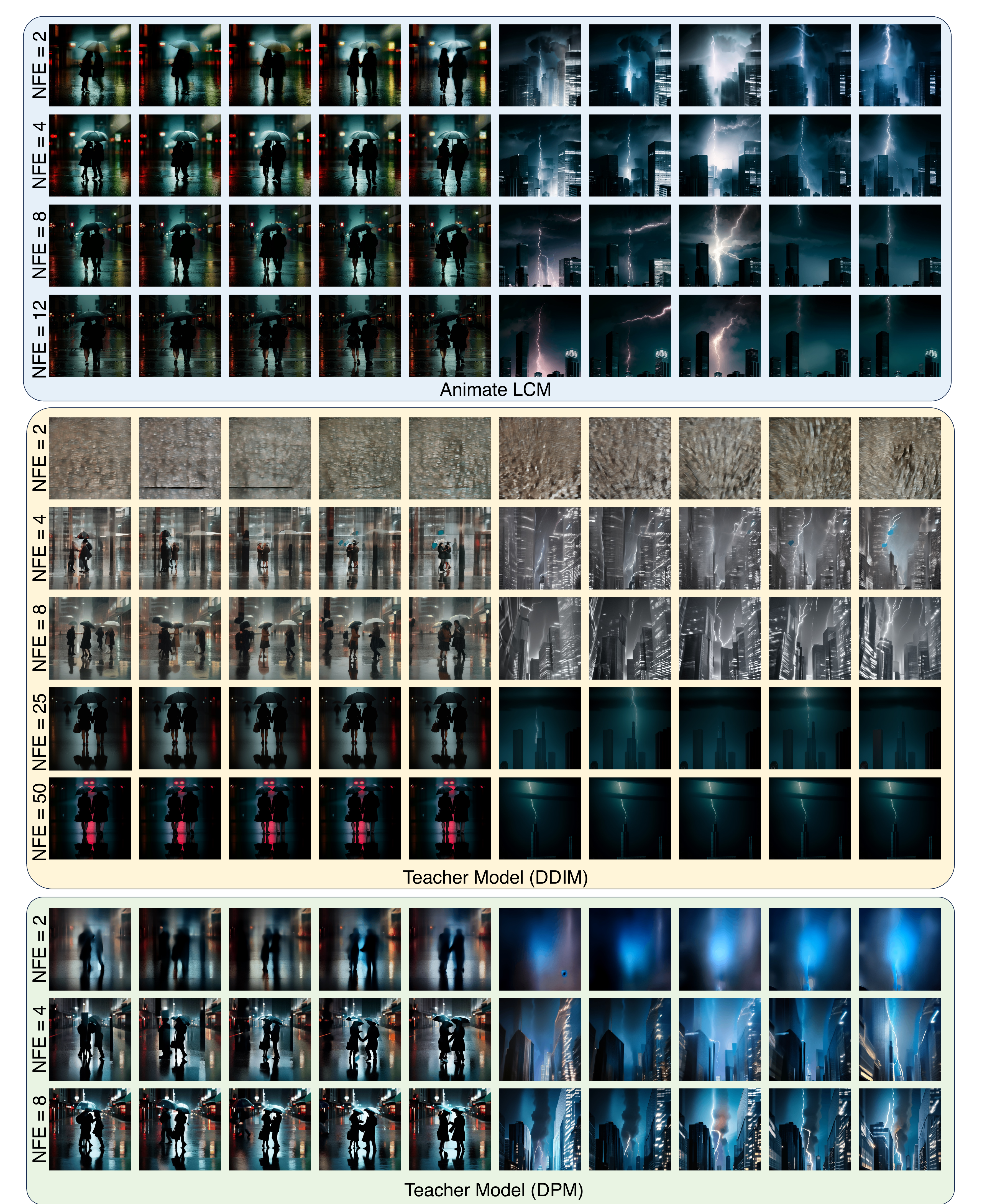}
\vspace{-1em}
    \caption{Qualitative comparison under different number of inference steps (NFE).}
    \label{fig:compare}
\end{figure}
 
\begin{figure}[t]
    \centering
\includegraphics[width=0.9\linewidth]{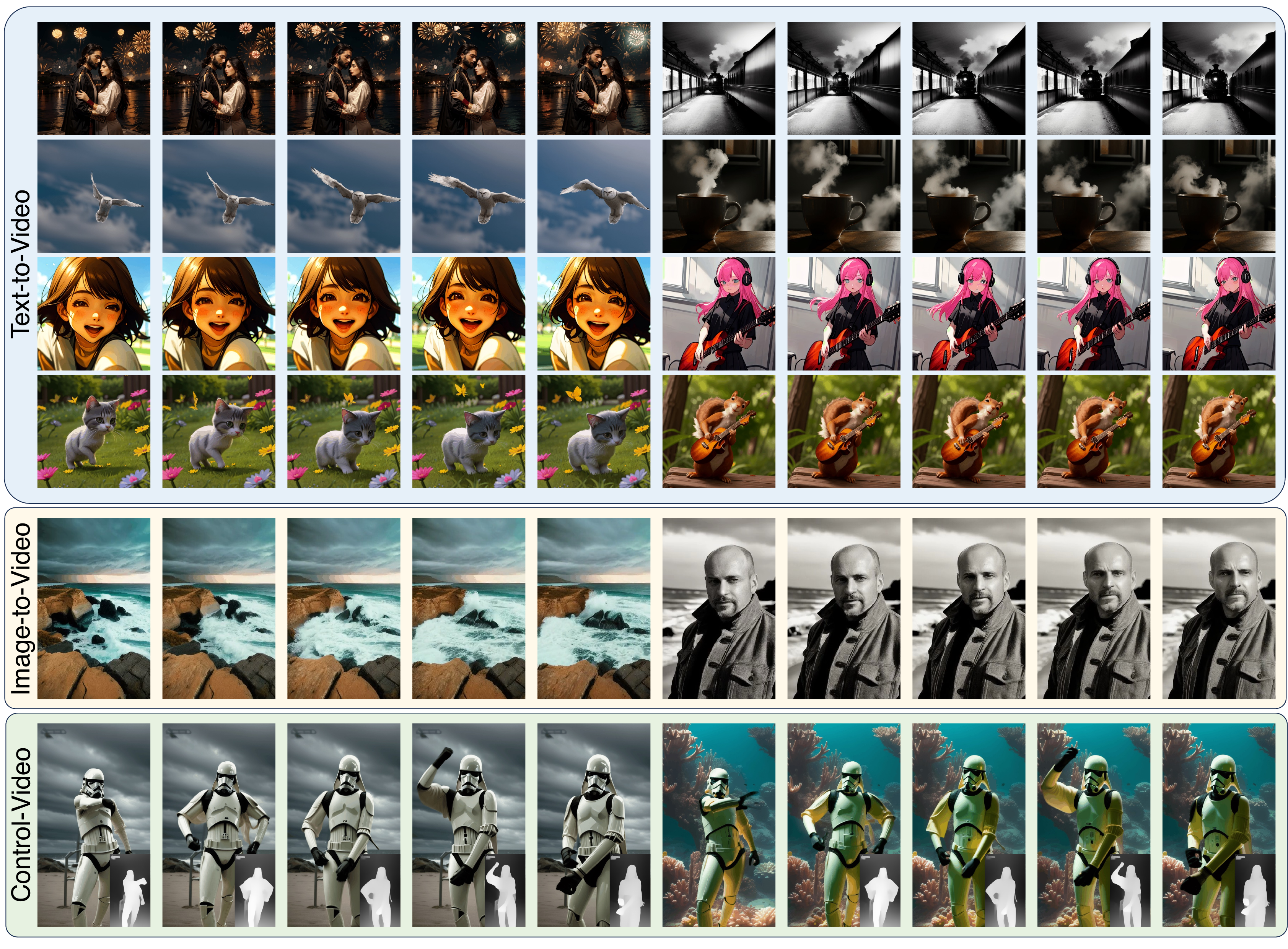}
\vspace{-1em}
    \caption{4-step generation results. AnimateLCM supports text-to-video, image-to-video, and controllable video generation.}
    \label{fig:demo}
\end{figure}

 \noindent \textbf{Qualitative results.} 
 Fig.~\ref{fig:demo} demonstrates the 4-step generation results of our method in text-conditioned video generation with different personalized style models including styles of realistic, 2D anime, and 3D anime, image-conditioned video generation, and layout-conditioned video generation.  We also demonstrate the generation results under different numbers of function evaluation~(NFEs) in Fig.~\ref{fig:compare}. We demonstrate good visual quality with only 2 inference steps. As the NFE increases, the generation quality increases accordingly, achieving competitive performance with the teacher model with 25 steps. 

\begin{table}[t]
    \centering
    \caption{Zero-shot video generation comparision on UCF-101. }~\label{tab:metric}
    \resizebox{0.9\linewidth}{!}{
    \begin{tabular}{l|cccc|cccc}
        \hline 
        \multirow{2}{*}{Methods} & \multicolumn{4}{c|}{FVD $\downarrow$} & \multicolumn{4}{c}{CLIPSIM $\uparrow$} \\
        \cline{2-9} 
        & 1-Step & 2-Step & 4-Step & 8-Step & 1-Step & 2-Step & 4-Step & 8-Step \\
        \hline
        DDIM~\cite{ddim}  & 4940.83 & 3218.74 & 1944.82 & 1209.88 & 4.43 & 5.26 & 14.87 & 24.38 \\
        DPM++~\cite{dpmsolver}  & 2731.37 & 2093.47 & 1043.82 & \uline{932.43} & 10.48 & 18.04 & 26.82 & 29.50 \\
        Ours & \uline{1256.50} & \uline{1081.26} & \textbf{925.71} & \textbf{910.34} & \uline{22.16} & \uline{25.99} & \uline{28.89} & \uline{30.03} \\
        Ours-R & \textbf{1071.50} & \textbf{790.99} & \uline{929.79} & 1081.72 & \textbf{25.41} & \textbf{29.39} & \textbf{30.62} & \textbf{30.71} \\
        \hline
    \end{tabular}}
    \vspace{-4mm}
\end{table}

\noindent \textbf{Quantitative Comparison.} Table ~\ref{tab:metric} illustrates quantitative metrics comparison for AnimateLCM and strong baseline methods DDIM~\cite{ddim}, and DPM++~\cite{dpmsolver}. AnimateLCM significantly surpasses the baseline methods, especially in the low step regime~(1$\sim$4). Additionally, all these metrics of AnimateLCM are evaluated without requiring classifier-free guidance~(CFG)~\cite{cfg} instead of 7.5 CFG strength applied for other baselines, thus saving half of the inference peak memory cost and inference time.  Additionally, we show Ours-R, which we replace the LDM weights with new weights finetuned on the high-quality personalized image datasets, can achieve even superior performance. It further indicates the effectiveness of our decouple learning approach.

\noindent \textbf{Advanced quantitative comparison.} For a more comprehensive evaluating of the ability of AnimateLCM, we apply Vbench~\cite{huang2024vbench} for a more advanced metric comparision, which including measurements from dozens of perspectives. We can observe from the Table.~\ref{tab:model_comparison} that, our model as the only video generation support fast generation (typically at lease 5 times faster than all compared methods), still achieves very competitive totoal score. 

\begin{table}[h!]
\centering
\vspace{-0.5em}
\caption{Advanced evaluation with Vbench.}
\vspace{-1em}\resizebox{0.8\linewidth}{!}{
\begin{tabular}{lcccc}
\toprule
Model Name       & Fast   & Total Score & Quality Score & Semantic Score \\
\midrule
Pika-1.0 (2024-06)           & N/A    & 80.69\%     & 82.92\%       & 71.77\%        \\
Gen-2 (2023-12)              & N/A    & 80.58\%     & 82.47\%       & 73.03\%        \\
VideoCrafter-1.0~\cite{he2022latent}             & $\times$   & 79.72\%     & 81.59\%       & 72.22\%        \\
AnimateLCM~(Ours)                   & $\checkmark$ & 79.42\%     & 82.36\%       & 67.65\%        \\
OpenSora V1.2                & $\times$   & 79.23\%     & 80.71\%       & 73.30\%        \\
Show-1~\cite{zhang2023show}                       & $\times$   & 78.93\%     & 80.42\%       & 72.98\%        \\
OpenSoraPlan V1.1            & $\times$   & 77.99\%     & 80.90\%       & 66.38\%        \\
AnimateDiff-V1~\cite{guo2023animatediff}               & $\times$   & 77.46\%     & 80.24\%       & 66.32\%        \\
Latte-1~\cite{ma2024latte}                      & $\times$   & 77.29\%     & 79.72\%       & 67.58\%        \\
Open-Sora~\cite{opensora}                    & $\times$   & 75.91\%     & 78.82\%       & 64.28\%        \\
\bottomrule
\end{tabular}}
\label{tab:model_comparison}
\vspace{-0.5em}
\end{table}
\noindent\textbf{Effectiveness of decoupled consistency learning.} We validate the effectiveness of our proposed decoupled distillation strategy.   For a fair comparison of convergence speed, we train the spatial LoRA weights for 4 hours on an 8 A100 GPUs. We then train our strategy on the video dataset for an additional 4 hours. We train the baseline without decoupled distillation for 8 hours.  Our strategy achieves FVD 985.9 and CLIPSIM 27.7 within the training budget while the baseline without the decoupled distillation strategy achieves FVD 1060.6 and CLIPSIM 18.8.

\noindent \textbf{Inference time comparision.} 
Diffusion models require 50 steps with proper CFG values for high-quality generation (50 $\times$ 2 model evaluations). Our model can generate videos in 4 steps without CFG (4 model evaluations).  Theoretically, our model can achieve acceleration by $\frac{50 \times 2}{4} = 25$  times. Testing on a single A800 with fp16 mixed precision, our model generates 2-second videos in 963ms, whereas normal diffusion models take 23564ms (24.47 times slower). Note that for the time computation, we exclude the VAE decoding time since it does not belong to the denoising process. 

\noindent \textbf{Denoising process visualization.} In Fig.~\ref{fig:denoise}, we visualize the denoising process of our model as well as that of a conventional video model. Within the same time frame, our model has generated three high-quality videos, while the compared video diffusion model has yet to complete the denoising of a single video.

\section{Conclusions and Limitations}
We present AnimateLCM, achieving computation-efficient personalized style video generation without personalized video data. Its decoupling strategies from two perspectives allows us to achieve fast stylized video generation with smaller training budget and alleviating the need to collect high-quality stylized video data. It might fail to generate samples with good quality with very low steps~(e.g., one-step) though.

\begin{acks}
This project is funded in part by National Key R\&D Program of China Project 2022ZD0161100, by the Centre for Perceptual and Interactive Intelligence (CPII) Ltd under the Innovation and Technology Commission (ITC)’s InnoHK, by General Research Fund of Hong Kong RGC Project 14204021. Hongsheng Li is a PI of CPII under the InnoHK. 
\end{acks}

\bibliographystyle{ACM-Reference-Format}
\bibliography{sample-base}


\begin{thebibliography}{23}


\ifx \showCODEN    \undefined \def \showCODEN     #1{\unskip}     \fi
\ifx \showDOI      \undefined \def \showDOI       #1{#1}\fi
\ifx \showISBNx    \undefined \def \showISBNx     #1{\unskip}     \fi
\ifx \showISBNxiii \undefined \def \showISBNxiii  #1{\unskip}     \fi
\ifx \showISSN     \undefined \def \showISSN      #1{\unskip}     \fi
\ifx \showLCCN     \undefined \def \showLCCN      #1{\unskip}     \fi
\ifx \shownote     \undefined \def \shownote      #1{#1}          \fi
\ifx \showarticletitle \undefined \def \showarticletitle #1{#1}   \fi
\ifx \showURL      \undefined \def \showURL       {\relax}        \fi
\providecommand\bibfield[2]{#2}
\providecommand\bibinfo[2]{#2}
\providecommand\natexlab[1]{#1}
\providecommand\showeprint[2][]{arXiv:#2}

\bibitem[Geng et~al\mbox{.}(2024)]%
        {geng2024consistency}
\bibfield{author}{\bibinfo{person}{Zhengyang Geng}, \bibinfo{person}{Ashwini Pokle}, \bibinfo{person}{William Luo}, \bibinfo{person}{Justin Lin}, {and} \bibinfo{person}{J~Zico Kolter}.} \bibinfo{year}{2024}\natexlab{}.
\newblock \showarticletitle{Consistency Models Made Easy}.
\newblock \bibinfo{journal}{\emph{arXiv preprint arXiv:2406.14548}} (\bibinfo{year}{2024}).
\newblock


\bibitem[Goodfellow et~al\mbox{.}(2014)]%
        {gan}
\bibfield{author}{\bibinfo{person}{Ian Goodfellow}, \bibinfo{person}{Jean Pouget-Abadie}, \bibinfo{person}{Mehdi Mirza}, \bibinfo{person}{Bing Xu}, \bibinfo{person}{David Warde-Farley}, \bibinfo{person}{Sherjil Ozair}, \bibinfo{person}{Aaron Courville}, {and} \bibinfo{person}{Yoshua Bengio}.} \bibinfo{year}{2014}\natexlab{}.
\newblock \showarticletitle{Generative adversarial networks}.
\newblock \bibinfo{journal}{\emph{NeurIPS}} (\bibinfo{year}{2014}).
\newblock


\bibitem[Guo et~al\mbox{.}(2023)]%
        {guo2023animatediff}
\bibfield{author}{\bibinfo{person}{Yuwei Guo}, \bibinfo{person}{Ceyuan Yang}, \bibinfo{person}{Anyi Rao}, \bibinfo{person}{Yaohui Wang}, \bibinfo{person}{Yu Qiao}, \bibinfo{person}{Dahua Lin}, {and} \bibinfo{person}{Bo Dai}.} \bibinfo{year}{2023}\natexlab{}.
\newblock \showarticletitle{Animatediff: Animate your personalized text-to-image diffusion models without specific tuning}.
\newblock \bibinfo{journal}{\emph{arXiv preprint arXiv:2307.04725}} (\bibinfo{year}{2023}).
\newblock


\bibitem[He et~al\mbox{.}(2022)]%
        {he2022latent}
\bibfield{author}{\bibinfo{person}{Yingqing He}, \bibinfo{person}{Tianyu Yang}, \bibinfo{person}{Yong Zhang}, \bibinfo{person}{Ying Shan}, {and} \bibinfo{person}{Qifeng Chen}.} \bibinfo{year}{2022}\natexlab{}.
\newblock \showarticletitle{Latent Video Diffusion Models for High-Fidelity Video Generation with Arbitrary Lengths}.
\newblock \bibinfo{journal}{\emph{arXiv preprint arXiv:2211.13221}} (\bibinfo{year}{2022}).
\newblock


\bibitem[Hessel et~al\mbox{.}(2021)]%
        {hessel2021clipscore}
\bibfield{author}{\bibinfo{person}{Jack Hessel}, \bibinfo{person}{Ari Holtzman}, \bibinfo{person}{Maxwell Forbes}, \bibinfo{person}{Ronan~Le Bras}, {and} \bibinfo{person}{Yejin Choi}.} \bibinfo{year}{2021}\natexlab{}.
\newblock \showarticletitle{Clipscore: A reference-free evaluation metric for image captioning}.
\newblock \bibinfo{journal}{\emph{arXiv preprint arXiv:2104.08718}} (\bibinfo{year}{2021}).
\newblock


\bibitem[Ho and Salimans(2022)]%
        {cfg}
\bibfield{author}{\bibinfo{person}{Jonathan Ho} {and} \bibinfo{person}{Tim Salimans}.} \bibinfo{year}{2022}\natexlab{}.
\newblock \showarticletitle{Classifier-free diffusion guidance}.
\newblock \bibinfo{journal}{\emph{arXiv preprint arXiv:2207.12598}} (\bibinfo{year}{2022}).
\newblock


\bibitem[Ho et~al\mbox{.}(2022)]%
        {ho2022video}
\bibfield{author}{\bibinfo{person}{Jonathan Ho}, \bibinfo{person}{Tim Salimans}, \bibinfo{person}{Alexey Gritsenko}, \bibinfo{person}{William Chan}, \bibinfo{person}{Mohammad Norouzi}, {and} \bibinfo{person}{David~J Fleet}.} \bibinfo{year}{2022}\natexlab{}.
\newblock \showarticletitle{Video diffusion models}.
\newblock \bibinfo{journal}{\emph{arXiv:2204.03458}} (\bibinfo{year}{2022}).
\newblock


\bibitem[Hu et~al\mbox{.}(2021)]%
        {hu2021lora}
\bibfield{author}{\bibinfo{person}{Edward~J Hu}, \bibinfo{person}{Yelong Shen}, \bibinfo{person}{Phillip Wallis}, \bibinfo{person}{Zeyuan Allen-Zhu}, \bibinfo{person}{Yuanzhi Li}, \bibinfo{person}{Shean Wang}, \bibinfo{person}{Lu Wang}, {and} \bibinfo{person}{Weizhu Chen}.} \bibinfo{year}{2021}\natexlab{}.
\newblock \showarticletitle{Lora: Low-rank adaptation of large language models}.
\newblock \bibinfo{journal}{\emph{arXiv preprint arXiv:2106.09685}} (\bibinfo{year}{2021}).
\newblock


\bibitem[Huang et~al\mbox{.}(2024)]%
        {huang2024vbench}
\bibfield{author}{\bibinfo{person}{Ziqi Huang}, \bibinfo{person}{Yinan He}, \bibinfo{person}{Jiashuo Yu}, \bibinfo{person}{Fan Zhang}, \bibinfo{person}{Chenyang Si}, \bibinfo{person}{Yuming Jiang}, \bibinfo{person}{Yuanhan Zhang}, \bibinfo{person}{Tianxing Wu}, \bibinfo{person}{Qingyang Jin}, \bibinfo{person}{Nattapol Chanpaisit}, {et~al\mbox{.}}} \bibinfo{year}{2024}\natexlab{}.
\newblock \showarticletitle{Vbench: Comprehensive benchmark suite for video generative models}. In \bibinfo{booktitle}{\emph{Proceedings of the IEEE/CVF Conference on Computer Vision and Pattern Recognition}}. \bibinfo{pages}{21807--21818}.
\newblock


\bibitem[Lu et~al\mbox{.}(2022)]%
        {dpmsolver}
\bibfield{author}{\bibinfo{person}{Cheng Lu}, \bibinfo{person}{Yuhao Zhou}, \bibinfo{person}{Fan Bao}, \bibinfo{person}{Jianfei Chen}, \bibinfo{person}{Chongxuan Li}, {and} \bibinfo{person}{Jun Zhu}.} \bibinfo{year}{2022}\natexlab{}.
\newblock \showarticletitle{Dpm-solver: A fast ode solver for diffusion probabilistic model sampling in around 10 steps}.
\newblock \bibinfo{journal}{\emph{Advances in Neural Information Processing Systems}}  \bibinfo{volume}{35} (\bibinfo{year}{2022}), \bibinfo{pages}{5775--5787}.
\newblock


\bibitem[Luo et~al\mbox{.}(2023)]%
        {lcm}
\bibfield{author}{\bibinfo{person}{Simian Luo}, \bibinfo{person}{Yiqin Tan}, \bibinfo{person}{Longbo Huang}, \bibinfo{person}{Jian Li}, {and} \bibinfo{person}{Hang Zhao}.} \bibinfo{year}{2023}\natexlab{}.
\newblock \showarticletitle{Latent consistency models: Synthesizing high-resolution images with few-step inference}.
\newblock \bibinfo{journal}{\emph{arXiv preprint arXiv:2310.04378}} (\bibinfo{year}{2023}).
\newblock


\bibitem[Ma et~al\mbox{.}(2024)]%
        {ma2024latte}
\bibfield{author}{\bibinfo{person}{Xin Ma}, \bibinfo{person}{Yaohui Wang}, \bibinfo{person}{Gengyun Jia}, \bibinfo{person}{Xinyuan Chen}, \bibinfo{person}{Ziwei Liu}, \bibinfo{person}{Yuan-Fang Li}, \bibinfo{person}{Cunjian Chen}, {and} \bibinfo{person}{Yu Qiao}.} \bibinfo{year}{2024}\natexlab{}.
\newblock \showarticletitle{Latte: Latent diffusion transformer for video generation}.
\newblock \bibinfo{journal}{\emph{arXiv preprint arXiv:2401.03048}} (\bibinfo{year}{2024}).
\newblock


\bibitem[Radford et~al\mbox{.}(2021)]%
        {clip}
\bibfield{author}{\bibinfo{person}{Alec Radford}, \bibinfo{person}{Jong~Wook Kim}, \bibinfo{person}{Chris Hallacy}, \bibinfo{person}{Aditya Ramesh}, \bibinfo{person}{Gabriel Goh}, \bibinfo{person}{Sandhini Agarwal}, \bibinfo{person}{Girish Sastry}, \bibinfo{person}{Amanda Askell}, \bibinfo{person}{Pamela Mishkin}, \bibinfo{person}{Jack Clark}, {et~al\mbox{.}}} \bibinfo{year}{2021}\natexlab{}.
\newblock \showarticletitle{Learning transferable visual models from natural language supervision}. In \bibinfo{booktitle}{\emph{ICML}}. PMLR, \bibinfo{pages}{8748--8763}.
\newblock


\bibitem[Shi et~al\mbox{.}(2024)]%
        {shi2024motion}
\bibfield{author}{\bibinfo{person}{Xiaoyu Shi}, \bibinfo{person}{Zhaoyang Huang}, \bibinfo{person}{Fu-Yun Wang}, \bibinfo{person}{Weikang Bian}, \bibinfo{person}{Dasong Li}, \bibinfo{person}{Yi Zhang}, \bibinfo{person}{Manyuan Zhang}, \bibinfo{person}{Ka~Chun Cheung}, \bibinfo{person}{Simon See}, \bibinfo{person}{Hongwei Qin}, {et~al\mbox{.}}} \bibinfo{year}{2024}\natexlab{}.
\newblock \showarticletitle{Motion-i2v: Consistent and controllable image-to-video generation with explicit motion modeling}. In \bibinfo{booktitle}{\emph{ACM SIGGRAPH 2024 Conference Papers}}. \bibinfo{pages}{1--11}.
\newblock


\bibitem[Singer et~al\mbox{.}(2022)]%
        {make-a-video}
\bibfield{author}{\bibinfo{person}{Uriel Singer}, \bibinfo{person}{Adam Polyak}, \bibinfo{person}{Thomas Hayes}, \bibinfo{person}{Xi Yin}, \bibinfo{person}{Jie An}, \bibinfo{person}{Songyang Zhang}, \bibinfo{person}{Qiyuan Hu}, \bibinfo{person}{Harry Yang}, \bibinfo{person}{Oron Ashual}, \bibinfo{person}{Oran Gafni}, {et~al\mbox{.}}} \bibinfo{year}{2022}\natexlab{}.
\newblock \showarticletitle{Make-a-video: Text-to-video generation without text-video data}.
\newblock \bibinfo{journal}{\emph{arXiv preprint arXiv:2209.14792}} (\bibinfo{year}{2022}).
\newblock


\bibitem[Song et~al\mbox{.}(2020)]%
        {ddim}
\bibfield{author}{\bibinfo{person}{Jiaming Song}, \bibinfo{person}{Chenlin Meng}, {and} \bibinfo{person}{Stefano Ermon}.} \bibinfo{year}{2020}\natexlab{}.
\newblock \showarticletitle{Denoising diffusion implicit models}.
\newblock \bibinfo{journal}{\emph{arXiv preprint arXiv:2010.02502}} (\bibinfo{year}{2020}).
\newblock


\bibitem[Song et~al\mbox{.}(2023)]%
        {cm}
\bibfield{author}{\bibinfo{person}{Yang Song}, \bibinfo{person}{Prafulla Dhariwal}, \bibinfo{person}{Mark Chen}, {and} \bibinfo{person}{Ilya Sutskever}.} \bibinfo{year}{2023}\natexlab{}.
\newblock \showarticletitle{Consistency models}.
\newblock \bibinfo{journal}{\emph{arXiv preprint arXiv:2303.01469}} (\bibinfo{year}{2023}).
\newblock


\bibitem[Soomro et~al\mbox{.}(2012)]%
        {soomro2012ucf101}
\bibfield{author}{\bibinfo{person}{Khurram Soomro}, \bibinfo{person}{Amir~Roshan Zamir}, {and} \bibinfo{person}{Mubarak Shah}.} \bibinfo{year}{2012}\natexlab{}.
\newblock \showarticletitle{UCF101: A dataset of 101 human actions classes from videos in the wild}.
\newblock \bibinfo{journal}{\emph{arXiv preprint arXiv:1212.0402}} (\bibinfo{year}{2012}).
\newblock


\bibitem[Unterthiner et~al\mbox{.}(2018)]%
        {fvd}
\bibfield{author}{\bibinfo{person}{Thomas Unterthiner}, \bibinfo{person}{Sjoerd Van~Steenkiste}, \bibinfo{person}{Karol Kurach}, \bibinfo{person}{Raphael Marinier}, \bibinfo{person}{Marcin Michalski}, {and} \bibinfo{person}{Sylvain Gelly}.} \bibinfo{year}{2018}\natexlab{}.
\newblock \showarticletitle{Towards accurate generative models of video: A new metric \& challenges}.
\newblock \bibinfo{journal}{\emph{arXiv preprint arXiv:1812.01717}} (\bibinfo{year}{2018}).
\newblock


\bibitem[Wang et~al\mbox{.}(2024)]%
        {wang2024phased}
\bibfield{author}{\bibinfo{person}{Fu-Yun Wang}, \bibinfo{person}{Zhaoyang Huang}, \bibinfo{person}{Alexander~William Bergman}, \bibinfo{person}{Dazhong Shen}, \bibinfo{person}{Peng Gao}, \bibinfo{person}{Michael Lingelbach}, \bibinfo{person}{Keqiang Sun}, \bibinfo{person}{Weikang Bian}, \bibinfo{person}{Guanglu Song}, \bibinfo{person}{Yu Liu}, {et~al\mbox{.}}} \bibinfo{year}{2024}\natexlab{}.
\newblock \showarticletitle{Phased Consistency Model}.
\newblock \bibinfo{journal}{\emph{arXiv preprint arXiv:2405.18407}} (\bibinfo{year}{2024}).
\newblock


\bibitem[Yu et~al\mbox{.}(2023)]%
        {yu2023magvit}
\bibfield{author}{\bibinfo{person}{Lijun Yu}, \bibinfo{person}{Yong Cheng}, \bibinfo{person}{Kihyuk Sohn}, \bibinfo{person}{Jos{\'e} Lezama}, \bibinfo{person}{Han Zhang}, \bibinfo{person}{Huiwen Chang}, \bibinfo{person}{Alexander~G Hauptmann}, \bibinfo{person}{Ming-Hsuan Yang}, \bibinfo{person}{Yuan Hao}, \bibinfo{person}{Irfan Essa}, {et~al\mbox{.}}} \bibinfo{year}{2023}\natexlab{}.
\newblock \showarticletitle{Magvit: Masked generative video transformer}. In \bibinfo{booktitle}{\emph{CVPR}}. \bibinfo{pages}{10459--10469}.
\newblock


\bibitem[Zhang et~al\mbox{.}(2023)]%
        {zhang2023show}
\bibfield{author}{\bibinfo{person}{David~Junhao Zhang}, \bibinfo{person}{Jay~Zhangjie Wu}, \bibinfo{person}{Jia-Wei Liu}, \bibinfo{person}{Rui Zhao}, \bibinfo{person}{Lingmin Ran}, \bibinfo{person}{Yuchao Gu}, \bibinfo{person}{Difei Gao}, {and} \bibinfo{person}{Mike~Zheng Shou}.} \bibinfo{year}{2023}\natexlab{}.
\newblock \showarticletitle{Show-1: Marrying pixel and latent diffusion models for text-to-video generation}.
\newblock \bibinfo{journal}{\emph{arXiv preprint arXiv:2309.15818}} (\bibinfo{year}{2023}).
\newblock


\bibitem[Zheng et~al\mbox{.}(2024)]%
        {opensora}
\bibfield{author}{\bibinfo{person}{Zangwei Zheng}, \bibinfo{person}{Xiangyu Peng}, \bibinfo{person}{Tianji Yang}, \bibinfo{person}{Chenhui Shen}, \bibinfo{person}{Shenggui Li}, \bibinfo{person}{Hongxin Liu}, \bibinfo{person}{Yukun Zhou}, \bibinfo{person}{Tianyi Li}, {and} \bibinfo{person}{Yang You}.} \bibinfo{year}{2024}\natexlab{}.
\newblock \bibinfo{booktitle}{\emph{Open-Sora: Democratizing Efficient Video Production for All}}.
\newblock
\urldef\tempurl%
\url{https://github.com/hpcaitech/Open-Sora}
\showURL{%
\tempurl}


\end{thebibliography}

\end{document}